# Robust Rigid Point Registration based on Convolution of Adaptive Gaussian Mixture Models

Can Pu[*], Nanbo Li[*], Robert B Fisher[*]

[*]University of Edinburgh
{canpu999, linanbo2008}@gmail.com      rbf@inf.ed.ac.uk

## Abstract

*Matching 3D rigid point clouds in complex environments robustly and accurately is still a core technique used in many applications. This paper proposes a new architecture combining error estimation from sample covariances and dual global probability alignment based on the convolution of adaptive Gaussian Mixture Models (GMM) from point clouds. Firstly, a novel adaptive GMM is defined using probability distributions from the corresponding points. Then rigid point cloud alignment is performed by maximizing the global probability from the convolution of dual adaptive GMMs in the whole 2D or 3D space, which can be efficiently optimized and has a large zone of accurate convergence. Thousands of trials have been conducted on 200 models from public 2D and 3D datasets to demonstrate superior robustness and accuracy in complex environments with unpredictable noise, outliers, occlusion, initial rotation, shape and missing points.*

## 1. Introduction

With the recent development of depth sensing devices and algorithms, 3D point clouds are more accessible to researchers. Integrating multi-modal 3D data under different noise, occlusion, outliers from multiple sensors or paths into a unified coordinate system robustly and accurately has become a core technique for various fields, such as 3D computer vision, reverse engineering, dimensional analysis, 3D modeling, robotics, virtual & augmented reality.

In real environments, noisy, incomplete, sometimes even wrong data affects previous registration algorithms [1,2,3,4] because of their underlying prior assumptions. Because of this, researchers have been investigating how to estimate the uncertainty of the acquired data for different sensors, such as the Kinect sensor [5], the time of flight sensor [6], the structure from motion sensor [7], the stereo vision sensor [8]. These suggest using resulting noise models for each point to represent their individual occurrence probability in 3D space. Thus, developing a new and robust rigid point cloud registration system that can exploit these error models has become a must.

In this paper, registering two point clouds is framed as maximizing the probability of the alignment of two adaptive Gaussian mixture models. As we know, no one has used it in the rigid point registration before. We will demonstrate that this approach is more robust and accurate compared with previous algorithms. The proposed dual adaptive GMMs (Gaussian Mixture Models) alignment can be optimized efficiently by the EM algorithm [9]. A new empirical approximation will be proposed to reduce the amount of calculation drastically.

The rest of this paper will be organized as the following. In Section 2, key previous registration algorithms will be reviewed briefly and also recent advances of the methods for estimating the uncertainty of the acquired data from different sensors. In Section 3, the rigid point registration theory is presented. In Section 4, 2D and 3D models from different scenes have been utilized to test the robustness and accuracy of our system in comparison with other key algorithms.

**Key contributions**: The key contributions of this paper are: 1) a new robust registration estimation algorithm based on maximizing probability over two probabilistic point clouds, 2) incorporating real 3D data covariance into the model, and 3) a novel approximation to the optimization step that greatly reduces computational complexity.

## 2. Literature review

In 1992, Besl and McKay [1] first introduced the Iterative Closest Point (ICP) algorithm to compute the rigid transformation between two point clouds by minimizing the Euclidean distance between the correponding points. Considering the probability distribution of each point, Segal et al. proposed Generalized-ICP [2] in 2009, which used the square Mahalanobis distance in essence to be the energy function and was based on correpondence searching using the same architecture with the standard ICP. There are many variants based on standard ICP such as different optimization (e.g. point-to-plane distance [10]) and so on.

Unlike ICP which regards the correspondence as binary, Robust Point Matching (e.g. Gold et al. [3]) utilizes soft correpondence, where the correpondence will range from zero to one although at the end they will converge to 1 or 0. In 2010, Myronenko and Song [4] proposed to treat one point cloud as a Gaussian mixture model and the other as the data points. Then they solve for the transformation that maximizes the probablity of the data points rather than the distance between correpondences. However, their Coherent Point Drift (CPD) algorithm assumed isotropic covariance for all the Gaussian functions, which does not accurately represent real 3D environments. There are many CPD variants as well, such as [11]. It chooses the optimal weight for the noise automatically in comparision to the original where the noise weight is manualy selected.

With the development of different depth sensors, more effective uncertainty estimation methods for various sensors have been designed. In 2012, Nguyen et al. [5] used distance and angle between the Kinect sensor and observed surface to estimate both axial and lateral noise distributions. In 2013, Engel et al. [7] used the geometric disparity error and photometric disparity error for the structure from motion sensor to estimate 3D point error. In 2015, Dal Mutto et al. [6] estimated the likelihood for the ToF (Time of Flight) sensor based on the physical properties of the sensor (eg: the IR frequency). In 2016, Marin et al. [8] developed an empirical uncertainty model for the stereo vision sensor based on the relationship between local cost and global cost.

In summary, there are previously developed methods for both robust rigid point cloud registration and modeling the estimation error of the points in a 3D point cloud. This paper improves registration accuracy and robustness using an approach that combines these two themes.

## 3. Methodology

### 3.1. The definition of adaptive GMM

The Gaussian function $g_{x_n}(\tau)$ of the point $x_n$ predicts the probablity that $x_n$ appears at the position $\tau$ in its own coordinate system. Based on Gaussian weights around each point, we will define a probability-like function that not only depends on the distribution of the point (represented by isotropic or anisotropic covariance) but also whether a possible corresponding point $cp_{x_n}$ in the other point cloud is nearby. We model the presence of a corresponding point by a weight function $w_{x_n}(\tau, cp_{x_n})$ that has significant value only when the corresponding point $cp_{x_n}$ is near the poistion $\tau$. A similar definition holds for the weight function $w_{y_m}(\tau, cp_{y_m})$.

In the analysis below, we assume that the **Y** point cloud has been already transformed from the initial point cloud $Y_0$ by rotation **R** and translation **t** (which then become the domain for the optimization of the evaluation function). The product $g_{x_n}(\tau) \times g_{y_m}(\tau)$ represents $x_n$, $y_n$ appearing at the same position $\tau$ in the same coordinate system. Thus, it encodes the underlying prior knowledge, that is, $x_n, y_n$ are possible corresponding points from two point clouds. In another word, any two points from fixed and moving point cloud can be the corresponding pair in our system and its likelihood should depend on the probability that $x_n$, $y_n$ appear at the same position $\tau$. It is obviously different from the ICP series [1,2,10], where the correspondence is binary.

Based on the previous analysis, we will design the adaptive GMM as follows, which we claim will achieve more accurate and robust estimation of the translation and rotation.

We define GMMs as:

$$G_X^{I,0}(\tau) = \sum_{n=1}^{N} w_{x_n}(\tau, cp_{x_n}) \times g_{x_n}(\tau)$$

$$G_Y^{R,t}(\tau) = \sum_{m=1}^{M} w_{y_m}(\tau, cp_{y_m}) \times g_{y_m}(\tau)$$

Where:

$$g_{x_n}(\tau) = \frac{1}{\sqrt{(2\pi)^D |\Sigma_{x_n}|}} e^{-\frac{1}{2}(\tau - x_n)^T \Sigma_{x_n}^{-1} (\tau - x_n)}$$

$$g_{y_m}(\tau) = \frac{1}{\sqrt{(2\pi)^D |\Sigma_{y_m}|}} e^{-\frac{1}{2}(\tau - y_m)^T \Sigma_{y_m}^{-1} (\tau - y_m)}$$

$$w_{x_n}(\tau, cp_{x_n}) = e^{-\frac{1}{2}(cp_{x_n} - \tau)^T \Sigma_{x_n}^{-1} (cp_{x_n} - \tau)}$$

$$w_{y_m}(\tau, cp_{y_m}) = e^{-\frac{1}{2}(cp_{y_m} - \tau)^T \Sigma_{y_m}^{-1} (cp_{y_m} - \tau)}$$

$G_X^{I,0}(\tau)$ denotes the GMM from the fixed point cloud X and $G_Y^{R,t}(\tau)$ represents the GMM from the moving point cloud Y after rotation $R$ and translation $t$. Thus $y_m = R * y_{m0} + t$, $\Sigma_{y_m} = R * \Sigma_{y_{m0}} * R'$, $|\Sigma_{y_m}| = |R * \Sigma_{y_{m0}} * R'| = |\Sigma_{y_{m0}}|$ due to $|R| = 1$. $\Sigma_{y_m}^{-1} = (R * \Sigma_{y_{m0}} * R')^{-1} = R * \Sigma_{y_{m0}}^{-1} * R'$ due to $R * R' = I$. In any time, $x_n = x_{n0}$ and $\Sigma_{x_n} = \Sigma_{x_{n0}}$.

### 3.2. The description of our model

By convolving the two adaptive GMMs and maximizing the marginal probability, we solve for a more accurate and robust solution for rigid point cloud matching. Fig 1 illustrates our model working on 2D data.

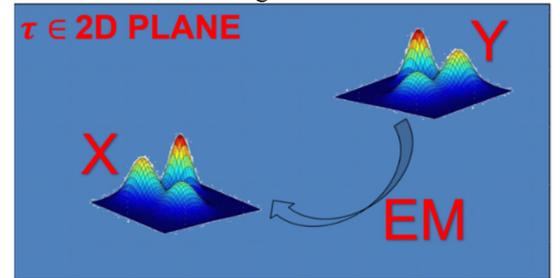

Figure 1: Demo on 2D data. X and Y are adaptive GMMs from the fixed and moving 2D point point cloud. $\tau$ is an arbitrary position on the 2D plane. The EM algorithm [9] will help Y move a bit in each iteration based on the cost from the convolution of X and Y on the whole 2D plane until convergence.

We now formulate the optimization over rotation R and translation t as an EM-like process. First, an energy function is defined as the following (where $\tau$ integrates over all the domain of the point clouds):

$$E = \int P(\tau) \times \log[G_X^{I,0}(\tau) \times G_Y^{R,t}(\tau)]\, d\tau \quad (1)$$

Where $P(\tau)$ is the probability that there is an point at the position $\tau$. It should be equal to the sum of the probability that all the possible correspoding pairs appear at the position $\tau$.

$$P(\tau) = \sum_{i=1}^{N} \sum_{j=1}^{M} P(\tau, x_i, y_j)$$

Where:
$$P(\tau, x_i, y_j) = \frac{1}{\sqrt{(2\pi)^D |\Sigma_{x_i}|}} e^{-\frac{1}{2}(\tau-x_i)^T \Sigma_{x_i}^{-1}(\tau-x_i)} \frac{1}{\sqrt{(2\pi)^D |\Sigma_{y_j}|}} e^{-\frac{1}{2}(\tau-y_j)^T \Sigma_{y_j}^{-1}(\tau-y_j)}$$

It represents the probablity that $x_i, y_j$ appears in the same position $\tau$.

Equation (1) can be rewritten as:

$$E = \int P(\tau) \times \log[\sum_{n=1}^{N} \sum_{m=1}^{M} F(\tau, y_m, x_n)]\, d\tau \quad (2)$$

Where:
$$F(\tau, y_m, x_n) = w_{x_n}(\tau, y_m) \times g_{x_n}(\tau) \times w_{y_m}(\tau, x_n) \times g_{y_m}(\tau)$$

We maximize this energy function to get the estimated rotation and translation. That is equal to minimizing its negative.

$$\min_{R,t}\left\{ \int P(\tau) \times (-\log[\sum_{n=1}^{N}\sum_{m=1}^{M} F(\tau, y_m, x_n)])\, d\tau \right\} \quad (3)$$

We adopt the EM algorithm [9,14] to solve for **R**, **t**. Its main idea is: guess "old" values of the parameters firstly and calculate a posteriori probablity $P^{old}(x_n, y_m|\tau)$ using Bayes' theorem then, which corresponds to expectation stage. After that, minimize the expectation of the negative log-likelihood function [14] to find the "new" parameters, which corresponds to maximization stage. Thus, we get:

$$\min_{R,t}\left\{ \left[\int \sum_{n=1}^{N}\sum_{m=1}^{M} P(\tau) P^{old}(x_n, y_m|\tau)(-\log[F^{new}(\tau, y_m, x_n)])\, d\tau\right]\right\} \quad (4)$$

Neglecting the constant term and using $P(\tau) \approx P^{old}(\tau)$, we simplify the target function:

$$\min_{R,t}\left\{ \left[\sum_{n=1}^{N}\sum_{m=1}^{M} \int P^{old}(\tau,\ x_n, y_m) \times \mathrm{Mal}^{new}(\tau,\ x_n, y_m)\, d\tau\right]\right\} \quad (5)$$

Where:
$$\mathrm{Mal}^{new}(\tau,\ x_n, y_m) = \frac{1}{2}(\tau - x_n)^T (\Sigma_{x_n}^{-1} + \Sigma_{y_m}^{-1})(\tau - x_n)$$
$$+ \frac{1}{2}(\tau - y_m)^T (\Sigma_{x_n}^{-1} + \Sigma_{y_m}^{-1})(\tau - y_m)$$

As we will justify below, there is no real benefit to integrate in the whole 2D or 3D space, because the values of the Gaussian functions are only significant at the data points themselves. Thus we need only evaluate each term $P^{old}(\tau,\ x_n, y_m) \times \mathrm{Mal}(\tau,\ x_n, y_m)$ essentially only at the positions of $x_n$ and $y_m$, which will reduce the time complexity to O[MN].

From another perspective, we want to minimize the energy function (5), if $x_n, y_m$ are corresponding points in the real situation, after our algorithm has converged, their Mahalanobis distance will be very small at $\tau = x_n$ and $y_m$ compared with the rest of the integration domain. If they are not the corresponding points, when $\tau = x_n$ or $y_m$ the Mahalanobis distance of the non-corresponding points becomes much bigger than the corresponding pairs. Thus we approximate the domain of the integral only using the positions of $x_n, y_m$.

Applying this simplification, the energy function becomes:

$$\min_{R,t}\left\{ \left[\sum_{n=1}^{N}\sum_{m=1}^{M} \sum_{\tau=x_n,y_m} P^{old}(\tau,\ x_n, y_m) \times \mathrm{Mal}^{new}(\tau,\ x_n, y_m)\, d\tau\right]\right\}$$

Expand the last sum and unite like terms:

$$\min_{R,t}\left\{ \left[\sum_{n=1}^{N}\sum_{m=1}^{M} C^{old}(m,n)(y_m - x_n)^T (\Sigma_{x_n}^{-1} + \Sigma_{y_m}^{-1})(y_m - x_n)\right]\right\} \quad (5)$$

Where:
$$C^{old}(m,n) = \frac{1}{\sqrt{(2\pi)^D |\Sigma_{x_n}{}^{old}|}} \frac{1}{\sqrt{(2\pi)^D |\Sigma_{y_m}{}^{old}|}} \left( e^{-\frac{1}{2}(y_m - x_n)^T \Sigma_{x_n}^{-1}(y_m - x_n)} + e^{-\frac{1}{2}(x_n - y_m)^T \Sigma_{y_m}^{-1}(x_n - y_m)} \right)^{old}$$
is coefficient.

We then minimize Equation (5) over the rotation **R** and translation **t** domain, using interior point optimization[15].

Additionally, we use the average minimum distance σ between two point clouds to control the covariance of Y point cloud in each iteration. We summarize our algorithm in the following diagram Fig. 2. $min(y_m, X)$ denotes the minimum distance between point $y_m$ and point cloud **X**.

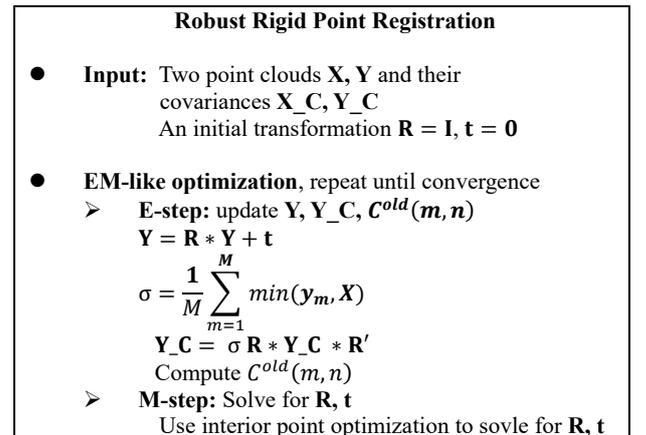

**Robust Rigid Point Registration**

- **Input:** Two point clouds **X, Y** and their covariances **X_C, Y_C**
  An initial transformation **R = I, t = 0**

- **EM-like optimization**, repeat until convergence
  > **E-step:** update Y, Y_C, $C^{old}(m,n)$
  $Y = R * Y + t$
  $$\sigma = \frac{1}{M}\sum_{m=1}^{M} min(y_m, X)$$
  $Y\_C = \sigma R * Y\_C * R'$
  Compute $C^{old}(m,n)$
  > **M-step:** Solve for **R, t**
  Use interior point optimization to sovle for **R, t**

Figure 2: Robust rigid point registration

## 4. Experiments

We implement our algorithm using matlab and Cuda C++. We run our algorithm on laptop Alienware 17, with Intel Core i7-7820HK processor (quad-core, 8MB cache, overclocking up to 4.4GHZ) and Nvidia Geforce GTX 1080 with 8GB GDDR5X. To test the accuracy and robustness of our algorithm, we divided all the experiments into two categories: 2D and 3D. On each part, our proposed method will be compared with coherent point drift [4] [1], point-to-plane ICP [10] using code from [12] [2], Generalized-ICP [2] [3] (only for 3D part). We tested the 2D part algorithm using the Gatorbait_100 database [4], using 100 fish with different shapes. The Stanford 3D Scanning Repository [5] and Trimbot2020 dataset [6] have been used for 100 3D models from multiple views of various objects and scenes in the real world. After that, a demonstration combining multiple kinect sensors' covariance estimation has been conducted to show the approach works much better in real surroundings compared with the other algorithms.

### 4.1. Experiment process

To synthesize the two point clouds to register, we randomly choose a model from the datasets above for two point clouds firstly. Then a different random long segment or a big patch of both models will be removed completely to simulate occlusion. After that, the two occluded models will be sampled differently to get their own point cloud respectively, which simulates the resolution of different sensors in real scenarios. Also, anisotropic Gaussian noise with random standard deviations and zero mean has been added to each point to simulate the complex noise in real environments resulting from known and unknown factors. Next, outliers have been added into the both point clouds to simulate noise points acquired by the sensors. Finally, an initial rigid transformation is applied to the moving point cloud.

All the experiments in either 2D or 3D are divided into four groups given 4 influence factors or variables: noise, outliers, occlusion, and initial rotation. In each group of experiments, one controlled variable will be changed and the values of the other variables will be picked randomly. The experiment has been conducted for 6 times at each controlled value, with a random shape and perturbation each time. We always prealign the two point clouds to zero mean and normalize them using the same scaling factor to make them robust to different rigid transformations. The covariance for each point will change with the rotation and the average minimum interval of all the points in the two point clouds in each iteration, which will make the whole system not easy to get to the local minimum.

---

[1] CPD code: https://sites.google.com/site/myronenko/research/cpd
[2] Point-to-Plane ICP code: http://www.cvlibs.net/software/libicp/
[3] G-ICP code: https://github.com/avsegal/gicp

We use the measure $\|I - R_{gt}R_{esti}^{-1}\|_F$ [13] to estimate the quality of the registration, where $R_{gt}$, $R_{esti}$ are the ground truth and estimated rotation matrix respectively and $\|\cdot\|_F$ is the Frobenius norm.

#### 4.1.1. Experiemnt results on 2D part

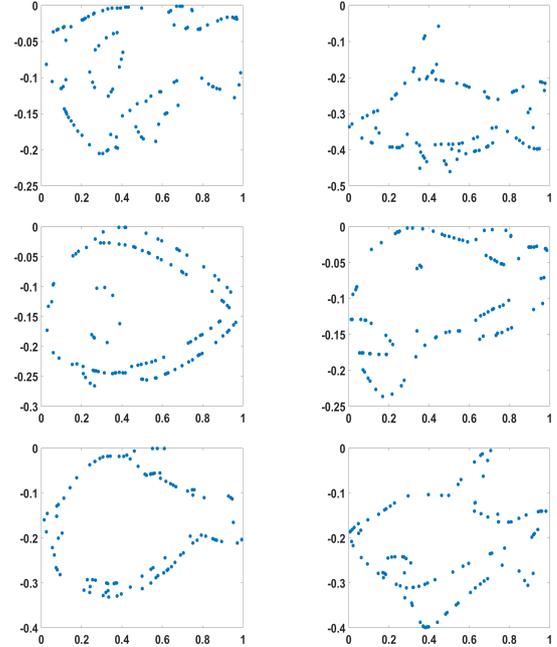

Figure 3: 2D models of various fish with different shapes.

We randomly extract about 100 points from each initial image as the 2D point cloud model. Fig 3 shows 6 examples with completely different shapes from the 100 images used to test the robustness to shape.

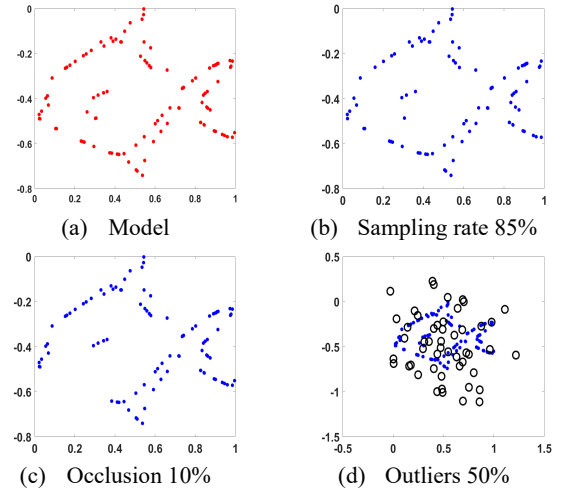

(a) Model    (b) Sampling rate 85%
(c) Occlusion 10%    (d) Outliers 50%

---

[4] Gatorbait_100 database: http://www.rvg.ua.es/graphs/dataset01.html
[5] Stanford Repository: http://graphics.stanford.edu/data/3Dscanrep/
[6] Trimbot2020 database: http://trimbot2020.webhosting.rug.nl/

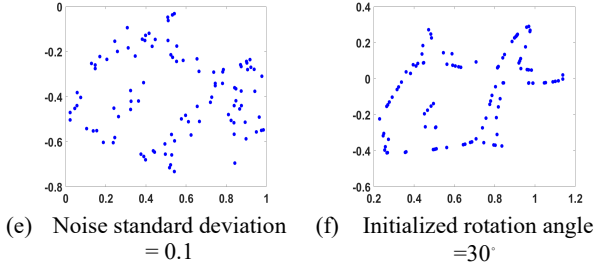

(e) Noise standard deviation = 0.1    (f) Initialized rotation angle =30°

Figure 4: Different influences from various factors.

Fig 4 illustrates the effect from the various factors which will be explored to assess the performance of the algorithm. In all 2D experiments, the sampling rate is set to 90% and 85% for fixed and moving point cloud, respectively. The remaining distribution factors are drawn from the uniform distribution from the Table 1 when they are not the controlled variable.

Table 1: Range for random factors.

| Initial rotation angle | [-15, 15] degree |
|---|---|
| Outliers | [0, 50] that is, [0, 50%] |
| Noise standard deviation | [0, 0.05] times radius of point cloud |
| Occlusion | [0, 0.05] that is: [0, 5%] |

Fig 5. shows a pair of point clouds before and after registration.

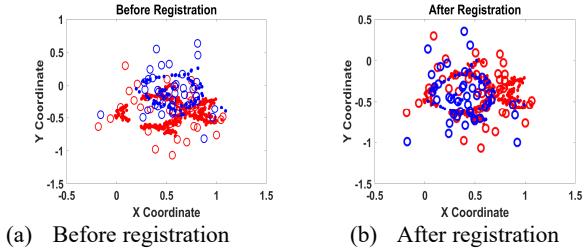

(a) Before registration    (b) After registration

Figure 5: The scene before and after registration.

When the rotation is the controlled variable, it will range from [-60, 60] degree and the step is equal to 8 degrees. Fig 6 shows the estimated error (standard deviation) with the change of the initial rotation. It is obvious that from [-36, 44] degrees our method is more accurate (closer to ground truth) and robust (smaller standard deviation) compared with coherent point drift and point-to-plane ICP. Beyond -36 or 44 degrees, our algorithm finds local minimum and breaks down.

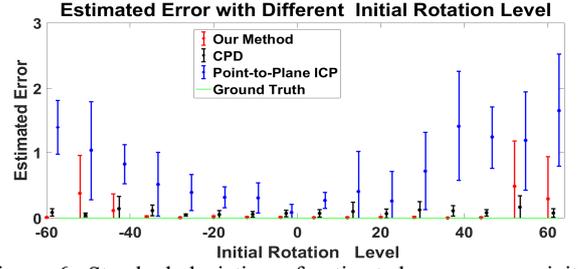

Figure 6: Standard deviation of estimated error versus initial rotations.

When noise level is the controlled variable, it will range from [0.01, 0.60] and its step is equal to 0.06. From Fig 7, we see with the increase of the noise level, our error increases slightly and is still very accurate and robust compared with CPD because the variances of all the noise on each axis have been stored in the covariances, which will be used by our system to estimate the error for each point.

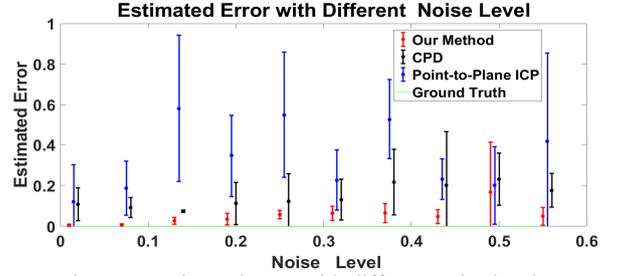

Figure 7: Estimated error with different noise levels.

When the number of outliers is the controlled variable, it ranges from [10, 200] (that is, from 10% to 200%) and the step is 20. We let the covariance for outliers be very big in both 2D and 3D part to represent they have a very low certainty. By doing that, the outliers will be filtered by our sysytem automatically. So, our experiment result looks much better than the comparison algorithms, see Fig 8.

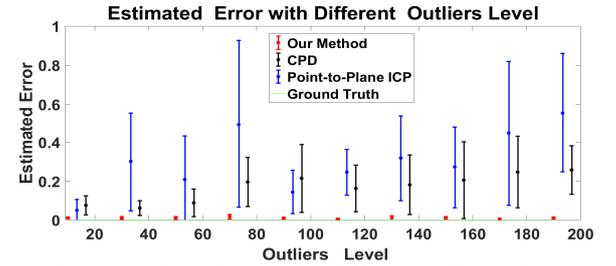

Figure 8: Estimated error with different outlier levels.

When occlusion is the controlled variable, it ranges from [0, 0.3] (from 0 to 30%) and the step size is equal to 0.03. The random occlusion part affects our model a lot when the occlusion rate is bigger than 15% because the missing long segments make it easy to converge to local minima, see Fig. 9.

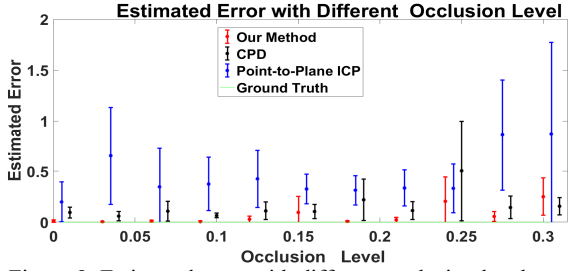

Figure 9: Estimated error with different occlusion levels.

In 2D part, we have tried 282 trials and the total time for each algorithm is listed in table 2. Point-to-Plane ICP used the minimum time mainly because it droped in the local minimum at first.

Table 2: Runing time in 2D part.

| Point-to-Plane ICP | 1.89 seconds |
|---|---|
| CPD | 27.94 seconds |
| Our methods (GTX 1080) | 622.52 seconds |

### 4.1.2. Experiment results on 3D part

From the Stanford 3D Scanning Repository and Trimbot2020 datasets 100 models were used, each with approximately 1000 points, from various views of different objects and scene. Fig. 10 shows 6 examples of the models used to compare algorithms.

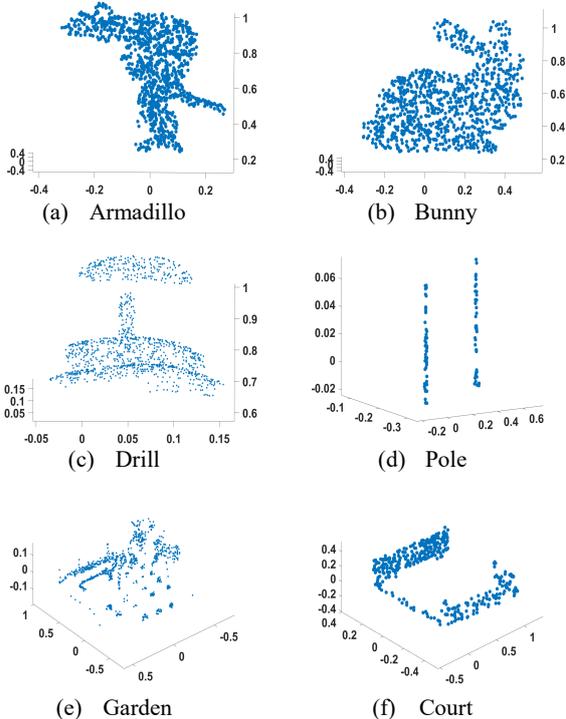

(a) Armadillo  (b) Bunny
(c) Drill  (d) Pole
(e) Garden  (f) Court

Figure 10: Different real 3D models with various views of multiple objects and scenes.

As with the 2D experiments, we apply different effects. Fig. 11 shows examples of the effects.

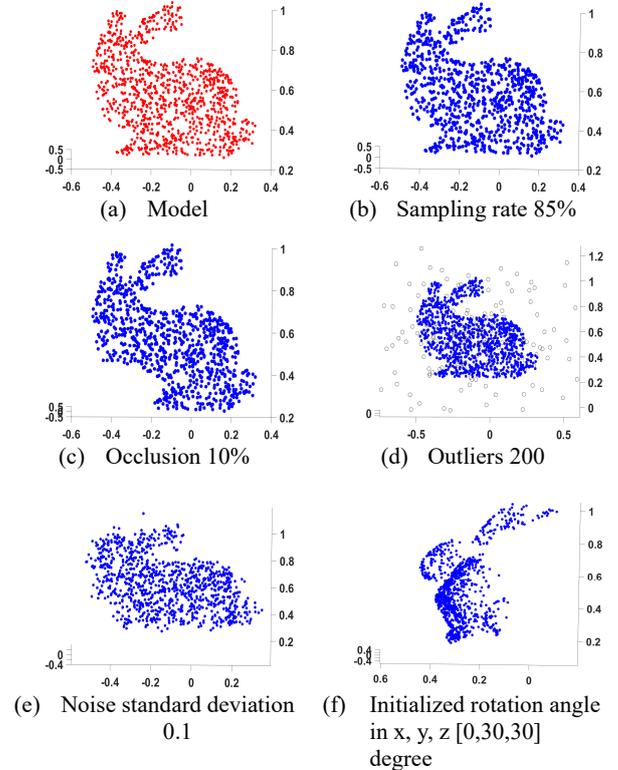

(a) Model  (b) Sampling rate 85%
(c) Occlusion 10%  (d) Outliers 200
(e) Noise standard deviation 0.1  (f) Initialized rotation angle in x, y, z [0,30,30] degree

Figure 11: Different influences from various factors.

Fig. 12 shows one pair of point clouds registered successfully. From the figure, we could see the hedges and trees overlap very well based on the maximum probability although there is a big patch of occlusion in both two point clouds, many outliers and noise.

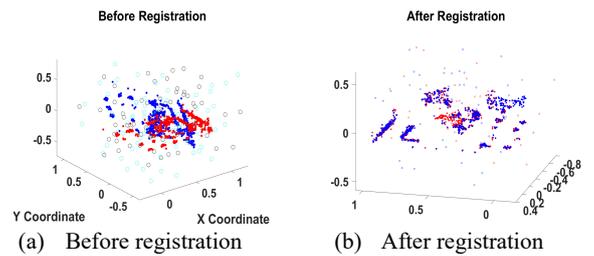

(a) Before registration  (b) After registration

Figure 12: A pair of successful registration.

In the four experiments below (ranging over rotation, noise, outliers, and occlusion), one controlled variable will vary and the other three variables will be chosen randomly. The experiment consists of 6 trials of each setting, where both the model and perturbation are chosen randomly. Table 3 gives specific information about the parameters:

Table 3: Range for random factors.

| Initial rotation angle around x, y, z axis | [-15, 15] degree |
|---|---|
| Outliers | [0, 500] that is, [0, 50%] |
| Noise standard deviation | [0, 0.01] times radius of point cloud |
| Occlusion | [0, 0.05] that is, [0, 5%] |

When initial rotation angle value is the controlled variable, it ranges from [-60,60], with an 8-degree step. In the experiements, the specific rotation angle around each axis is chosen as 0 or the initial rotation angle value randomly. Fig 13 shows the performance with the change of initial rotation angle value.

Beyond -44 or 36 degrees, our algorithm breaks down but within [-44,36] degrees, our algorithm is much more stable and accurate compared with the other algorithms.

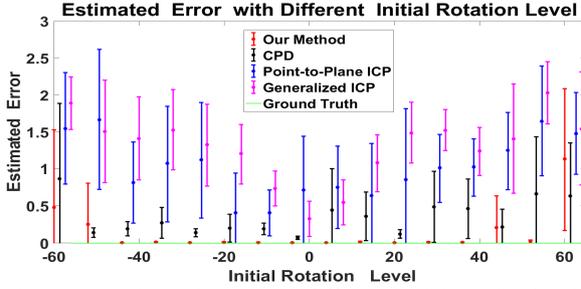
Figure 13: Estimated error with different initial rotation levels.

When noise level is the controlled variable, it ranges from [0.01, 0.6], with a step equal to 0.06. Fig 14 shows robust and accurate performace compared with the rest.

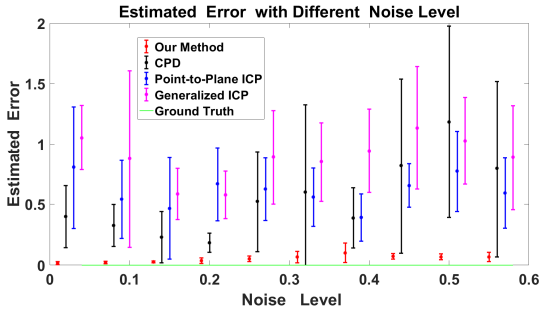
Figure14: Estimated error with different noise levels.

When outliers is the controlled variable, it ranges from [100, 2000] (from 10% to 200%), with a step equal to 200. The covariance for each outlier will be very big like that in the 2D part. Fig 15 shows the proposed algorithm has superior performance again.

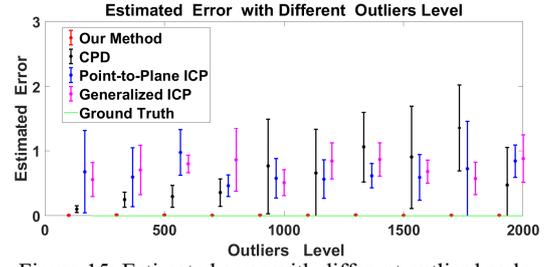
Figure 15: Estimated error with different outlier levels.

When occlusion rate is the controlled variable, it is chosen from [0,0.3] (from 0 to 30%), with a step equal to 0.03. Fig. 16 shows that within 15%, the proposed algorithm performs very well. If the occlusion exceeds 15%, the algorithm converges to a local minimum.

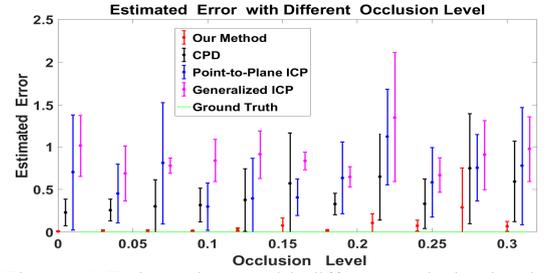
Figure 16: Estimated error with different occlusion levels.

In 3D part, we have tried 282 trials and the total time for each algorithm in 3D part listed in table 4.

Table 4: Running time in 3D part.

| Point-to-Plane ICP | 29.81 seconds |
|---|---|
| Generalized ICP | 274.36 seconds |
| CPD | 429.52 seconds |
| Our method (GTX 1080) | 2820.32 seconds |

### 4.2. Real data from multiple Kinect Sensors

One obvious difference between our algorithm and the previous methods is that we require the covariance of each point as input. There are many 3D point covariance estimation methods [eg: 5,6,7,8] for different kinds of sensors.

Based on the concept in [5] we propose a simple but general covariance estimation method for Kinect sensors. The uncertainty of each valid 3D point acquired by the Kinect sensor will depend on the depth value d and the angle $\alpha$ between the camera and the normal of the surface.

$$U(\alpha, d) = \exp[w_1(1 - \cos\alpha) + w_2 d] \quad (6)$$

Equation (6) shows that, with the increase of the depth d and angle $\alpha$, the uncertainty U of the corresponding 3D point will increase. We get $w_1 = 1.6658$ and $w_2 = 0.2776$ by letting $U\left(\frac{\pi}{3}, 0\right) = U(0,3) = 2.2$. The number 2.2 is set manually and the algorithm works well if that number is in the range [1,10] (known by our experiements). Then we multiplied the uncertainty and the identity matrix to estimate the covariance for each point.

We tested our algorithm using data from two fixed kinect sensors. The ground truth of the rotation between the two kinect sensors is known by calibration. Fig. 17 (a) (b) show the scene before and after registration using our algorithm. Fig. 17 (c) adds the color texture information into the two registered point clouds.

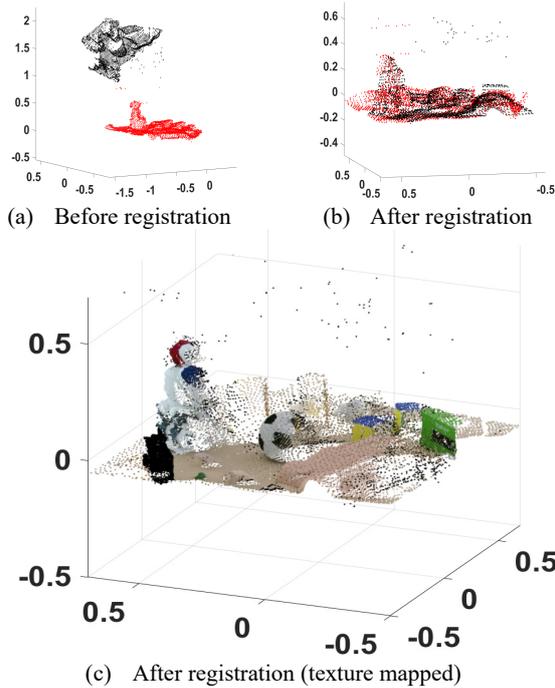

(a) Before registration  (b) After registration

(c) After registration (texture mapped)

Figure 17: (a) the scene before registration, (b) After registration, (c) The registered two point clouds have been added color into.

In the experiment, the two point clouds acquired by the two kinect sensors have been downsampled to 3139 and 4029 points using the Grid average method. Then we provided the same initial rotation to the four algorithms. After all the algorithms have converged, the estimated error of our algorithm is lowest (0.1202), see Fig. 18.

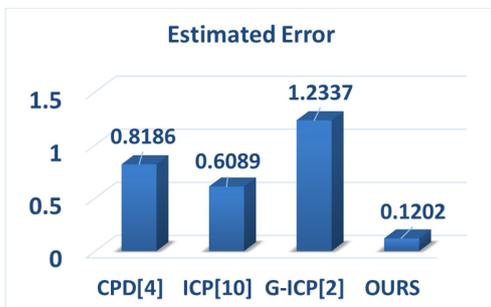

Figure 18: Estimated error.

## 5. Conclusion

The proposed robust rigid point registration algorithm is based on convolution of adaptive Gaussian mixture models. It provides a new architecture for accurate and robust rigid point matching. Considering the vast computation requirement, we used an approximation to reduce the time complexity from $\mathbb{O}[MN(M+N)]$ to $\mathbb{O}[MN]$. In the 2D experiments, we also explored a range of point densities, but did not find any obvious evidence to show the proposed algorithm will be sensitive to the density of the data. All the experiments in section 4 show that the proposed method is very robust and accurate and works well. The obvious difference between our algorithm and the previous ones is that it needs covariances at each point as input, which requires error models of how to estimate the real covariance for each kind of sensor. The code for the proposed algorithm can be downloaded from: *https://github.com/Canpu999/Robust-Rigid-Point-Registration*.

## Acknowlegments

This project received funding from the the European Union's Horizon 2020 research and innovation programme under grant No. 688007 (Trimbot2020).

## References

[1] P. J. Besl and N. D. McKay. A method for registration of 3-D shapes. *IEEE Transactions on Pattern Analysis and Machine Intelligence*, 14(2), pages 239–256, 1992

[2] A. Segal, D. Haehnel, and S. Thrun. Generalized-ICP. In *Robotics: Science and Systems*, page 435, 2009.

[3] S. Gold, A. Rangarajan, C.-P. Lu, S. Pappu, and E. Mjolsness. New algorithms for 2D and 3D point matching. *Pattern Recognition*, 31(8): 1019–1031, 1998.

[4] A. Myronenko and Xubo Song. Point Set Registration: Coherent Point Drift. *IEEE Transactions on Pattern Analysis and Machine Intelligence*, 32(12): 2262–2275, 2010.

[5] C. V. Nguyen, S. Izadi, and D. Lovell. Modeling Kinect Sensor Noise for Improved 3D Reconstruction and Tracking. In *2012 Second International Conference on 3D Imaging, Modeling, Processing, Visualization & Transmission*, pages 524–530, 2012.

[6] C. D. Mutto, P. Zanuttigh, and G. M. Cortelazzo. Probabilistic ToF and Stereo Data Fusion Based on Mixed Pixels Measurement Models. *IEEE Transactions on Pattern Analysis and Machine Intelligence*, 37(11): 2260–2272, 2015.

[7] J. Engel, J. Sturm, and D. Cremers. Semi-dense Visual Odometry for a Monocular Camera. In *2013 IEEE International Conference on Computer Vision*, pages 1449–1456, 2013.

[8] G. Marin, P. Zanuttigh, and S. Mattoccia. Reliable Fusion of ToF and Stereo Depth Driven by Confidence Measures. In *2016 European Conference on Computer Vision*, pages 386–401, 2016.

[9] A. P. Dempster, N. M. Laird, and D. B. Rubin. Maximum Likelihood from Incomplete Data via the EM Algorithm.

*Journal of the Royal Statistical Society. Series B (methodological)*, 39(1):1–38, 1977.

[10] Y. Chen and G. Medioni. Object modelling by registration of multiple range images. *Image and Vision Computing*, 10(3): 145–155, 1992.

[11] P. Wang, P. Wang, Z. Qu, Y. Gao, and Z. Shen. A refined coherent point drift (CPD) algorithm for point set registration. *Science China Information Sciences*, 54(12): 2639–2646, 2011.

[12] A. Geiger, P. Lenz, and R. Urtasun. Are we ready for autonomous driving? The KITTI vision benchmark suite. In *2012 IEEE Conference on Computer Vision and Pattern Recognition*, pages 3354–3361, 2012.

[13] D. Q. Huynh. Metrics for 3D Rotations: Comparison and Analysis. *Journal of Mathematical Imaging and Vision*, 35(2): 155–164, 2009.

[14] C. M. Bishop. *Neural Networks for Pattern Recognition*. New York, NY, USA: Oxford University Press, Inc., 1995.

[15] S. J. Wright. Primal-dual interior-point methods. In *Society for Industrial and Applied Mathematics*, 1997.